\documentclass[11pt,a4paper]{article}
\usepackage[hyperref]{acl2019}
\usepackage{times}
\usepackage{latexsym}

\usepackage{url}

\aclfinalcopy 
\def\aclpaperid{62} 

\usepackage[utf8]{inputenc}
\usepackage{amssymb}
\usepackage{amsmath}
\usepackage{algorithm}
\usepackage[noend]{algpseudocode}

\makeatletter
\def\BState{\State\hskip-\ALG@thistlm}
\makeatother

\usepackage{bm}
\usepackage{multirow}
\usepackage{tablefootnote}

\usepackage{float}
\usepackage{subcaption}
\usepackage{mathtools, nccmath}
\usepackage{times}
\usepackage{latexsym}
\usepackage{array}
\usepackage{url}
\usepackage{acl2019}

\newcommand{\hist}{\mbox{$\mathbf{y}_{1:i-1}$}}

\title{UCAM Biomedical translation at WMT19: Transfer learning multi-domain ensembles}
  
\author{Danielle Saunders$^\dag$ \and Felix Stahlberg$^\dag$ \and Bill Byrne$^{\ddagger\dag}$ \\
\\
    $^\dag$Department of Engineering, University of Cambridge, UK  \\
\\
    $^\ddagger$SDL Research, Cambridge, UK}

\begin{document}

\allowdisplaybreaks
\maketitle
\begin{abstract}
The 2019 WMT Biomedical translation task involved translating Medline abstracts. We approached this using transfer learning to obtain a series of strong neural models on distinct domains, and combining them into multi-domain ensembles. We further experiment with an adaptive language-model ensemble weighting scheme. Our submission achieved the best submitted results on both directions of English-Spanish.  
\end{abstract}

\section{Introduction}
Neural Machine Translation (NMT) in the biomedical domain presents challenges in addition to general domain translation. Firstly, available corpora are relatively small, exacerbating the effect of noisy or poorly aligned training data. Secondly, individual sentences within a biomedical document may use specialist vocabulary from small domains like health or statistics, or may contain generic language. Training to convergence on a single biomedical dataset may therefore not correspond to good performance on arbitrary biomedical test data.

Transfer learning is an approach in which a model is trained using knowledge from an existing model \cite{khan2018hunter}. Transfer learning typically involves initial training on a large, general domain corpus, followed by fine-tuning on the domain of interest. We apply transfer learning iteratively on datasets from different domains, obtaining strong models that cover two domains for both directions of the English-German language pair, and three domains for both directions of English-Spanish.

The domain of individual documents in the 2019 Medline test dataset is unknown, and may vary sentence-to-sentence. Evenly-weighted ensembles of models from different domains can give good results in this case \cite{freitag2016fast}. However, we suggest a better approach would take into account the likely domain, or domains, of each test sentence. We therefore investigate applying Bayesian Interpolation for language-model based multi-domain ensemble weighting.

\subsection{Iterative transfer learning}
Transfer learning has been used to adapt models both across domains, e.g. news to biomedical domain adaptation, and within one domain, e.g. WMT14 biomedical data to WMT18 biomedical data \cite{khan2018hunter}. For en2de and de2en we have only one distinct in-domain dataset, and so we use standard transfer learning from a general domain news model.

For es2en and en2es, we use the domain-labelled Scielo dataset to provide two distinct domains, health and biological sciences (`bio'), in addition to the complete biomedical dataset \cite{neves2016scielo}. We therefore experiment with iterative transfer learning, in which a model trained with transfer learning is then trained further on the original domain. 

NMT transfer learning for domain adaptation involves using the performance of a model on some general domain $A$ to improve performance on some other domain $B$: $A\rightarrow B$. However, if the two domains are sufficiently related, we suggest that task $B$ could equally be used for transfer learning $A$: $B\rightarrow A$. The stronger general model $A$ could then be used to achieve even better performance on other tasks: $B\rightarrow A \rightarrow B$,  $B\rightarrow A \rightarrow C$, and so on.

\subsection{Adaptive decoding}
Previous work on transfer learning typically aims to find a single model that performs well on a known domain of interest \cite{khan2018hunter}. The biomedical translation task offers a scenario in which the test domain is unknown, since individual Medline documents can have very different styles and topics. Our approach is to decode such test data with an ensemble of distinct domains.

For intuitive ensemble weights, we use sequence-to-sequence Bayesian Interpolation (BI) as described in \citet{saunders2019domain}, which also contains a more in-depth derivation and discusses possible hyperparameter configurations. We consider models $p_k(\mathbf{y} | \mathbf{x})$ trained on $K$ domains, used for $T=K$ domain decoding tasks. We assume throughout that $p(t)=\frac{1}{T}$, i.e.\ that tasks are equally likely absent any other information. Weights $\lambda_{k,t}$ define a task-conditional ensemble. At step $i$, where $h_i=\hist$ is decoding history:
\begin{align}
 \label{eq:adaptiveensemble}
         p(y_i|h_i,\mathbf{x}) = \sum_{k=1}^K  p_k(y_i|h_i, \mathbf{x})  \sum_{t=1}^T p(t|h_i, \mathbf{x}) \lambda_{k,t} 
 \end{align}
This is an adaptively weighted ensemble where, for each source sentence $\mathbf{x}$ and output hypothesis $\mathbf{y}$, we re-estimate $p(t|h_i,\mathbf{x})$ at each step:

\begin{equation}p (t|h_i, \mathbf{x}) = \frac{p(h_i|t,\mathbf{x}) p(t|\mathbf{x})} {\sum_{t'=1}^T p(h_i|t',\mathbf{x}) p(t'|\mathbf{x})}\label{eq:taskpost}\end{equation}
$p(h_i|t,\mathbf{x})$ is found from the last score of each model:
\begin{align} 
p(h_i|t,\mathbf{x}) &= p(y_{i-1}|h_{i-1}, t,\mathbf{x}) \\ \notag
&= \sum_k p_k(y_{i-1}|h_{i-1}, t,\mathbf{x}) \lambda_{k,t} 
\end{align}
We use $G_t$, an n-gram language model trained on source training sentences from task $t$, to estimate initial task posterior $p( t | \mathbf{x})$:
\begin{align} 
 \label{eq:lmsentence}
\frac{p(\mathbf{x}|t)p(t)}{\sum_{t'=1}^Tp(\mathbf{x}|t')p(t')} =\frac{G_t(\mathbf{x})^\alpha}{\sum_{t'=1}^T G_{t'}(\mathbf{x})^\alpha} 
\end{align}

Here $\alpha$ is a smoothing parameter. If $\overline{G}_{k,t} = \sum_{\mathbf{x} \in \text{Test}_t} G_k(\mathbf{x})$, we take: \begin{equation}
    \lambda_{k,t} = \frac{\overline{G}_{k,t}^\alpha}{\sum_{k'} \overline{G}_{k',t}^\alpha}
    \label{eq:lambda-bisd}
\end{equation} 

\begin{figure}[t!]
\centering
\small
\includegraphics[width=\linewidth]{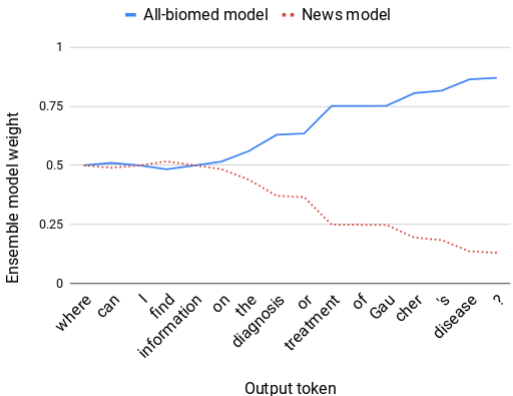}
\caption{Adaptively adjusting model weights during decoding with Bayesian Interpolation}
\label{fig:bi}
\end{figure}
Figure \ref{fig:bi} demonstrates this adaptive decoding when weighting a biomedical and a general (news) domain model to produce a biomedical sentence. The model weights are even until biomedical-specific vocabulary is produced, at which point the in-domain model dominates.
\begin{table*}[t]
\centering
\small
\begin{tabular}{|l|l|l|l|l|l|}\hline
& \textbf{Domain}  & \textbf{Training datasets} & \textbf{Sentence pairs} & \textbf{Dev datasets} & \textbf{Sentence pairs}\\ \hline
  \multirow{9}{*}{es-en}& \multirow{5}{*}{All-biomed}& UFAL Medical\footnotemark &639K & \multirow{5}{*}{Khresmoi\footnotemark} & \multirow{5}{*}{1.5K} \\
 & &Scielo\footnotemark & 713K &&  \\
 & & Medline titles\footnotemark& 288K&&  \\ 
 & & Medline training abstracts & 83K && \\
 & & Total (pre) / post-filtering &(1723K) / \textbf{1291K}&& \\
  \cline{2-6}
& \multirow{2}{*}{Health} & Scielo health only & 587K & \multirow{2}{*}{Scielo 2016 health}& \multirow{2}{*}{5K}\\

 & &Total post-filtering &\textbf{558K}&&  \\
 \cline{2-6}
& \multirow{2}{*}{Bio} & Scielo bio only & 126K &\multirow{2}{*}{Scielo 2016 bio}& \multirow{2}{*}{4K} \\
 & &Total post-filtering &\textbf{122K} && \\

 \hline
 \multirow{3}{*}{en-de} &  \multirow{3}{*}{All-biomed} & UFAL Medical  &2958K& Khresmoi&  1.5K\\
 && Medline training abstracts & 33K &Cochrane\footnotemark &467 \\
  & & Total (pre) / post-filtering &(2991K) / \textbf{2156K}& & \\
 \hline

\end{tabular}
\caption{Biomedical training and validation data used in the evaluation task. For both language pairs identical data was used in both directions. }\label{tab:data}
\end{table*}

\footnotetext[1]{\url{https://ufal.mff.cuni.cz/ufal\_medical\_corpus}} 
\footnotetext[2]{\citet{dusek2017khresmoi}}
\footnotetext[3]{\citet{neves2016scielo}}
\footnotetext[4]{\url{https://github.com/biomedical-translation-corpora/medline} \cite{yepes2017findings}}
\footnotetext[5]{\url{http://www.himl.eu/test-sets}}

\subsection{Related work}
Transfer learning has been applied to NMT in many forms. \citet{luong2015stanford} use transfer learning to adapt a general model to in-domain data. \citet{zoph2016transfer} use multilingual transfer learning to improve NMT for low-resource languages. \citet{chu2017empirical} introduce mixed fine-tuning, which carries out transfer learning to a new domain combined with some original domain data. \citet{kobus2017domain} train a single model on multiple domains using domain tags. 
\citet{khan2018hunter} sequentially adapt across multiple biomedical domains to obtain one single-domain model.

At inference time, \citet{freitag2016fast} use uniform ensembles of general and fine-tuned models. Our Bayesian Interpolation experiments extend previous work by \citet{allauzen2011bayesian} on Bayesian Interpolation for language model combination.

\section{Experimental setup}
\label{ss:data}
\subsection{Data}
We report on two language pairs: Spanish-English (es-en) and English-German (en-de). Table \ref{tab:data} lists the data used to train our biomedical domain evaluation systems. For en2de and de2en we additionally reuse strong general domain models trained on the WMT19 news data, including filtered Paracrawl. Details of data preparation and filtering for these models are discussed in \citet{ucam-wmt19}.

For each language pair we use the same training data in both directions, and use a 32K-merge source-target BPE vocabulary \cite{sennrich2016subword} trained on the `base' domain training data (news for en-de, Scielo health for es-en)

For the biomedical data, we preprocess the data using Moses tokenization, punctuation normalization and truecasing. We then use a series of simple heuristics to filter the parallel datasets:
\begin{itemize}
    \item Detected language filtering using the Python \texttt{langdetect} package\footnote{\url{https://pypi.org/project/langdetect/}}. In addition to mislabelled sentences, this step removes many sentences which are very short or have a high proportion of punctuation or HTML tags.
    \item Remove sentences containing more than 120 tokens or fewer than 3.
    \item Remove duplicate sentence pairs
    \item Remove sentences where the ratio of source to target tokens is less than 1:3.5 or more than 3.5:1
    \item Remove pairs where more than 30\% of either sentence is the same token. 
\end{itemize}

\subsection{Model hyperparameters and training}
We use the Tensor2Tensor implementation of the Transformer model with the \texttt{transformer\_big} setup for all NMT models \cite{tensor2tensor}. By default this model size limits batch size of 2K due to memory constraints. We delay gradient updates by a factor of 8, letting us effectively use a 16K batch size  \cite{saunders2018multi}. We train each domain model until it fails to improve on the domain validation set in 3 consecutive checkpoints, and perform checkpoint averaging over the final 10 checkpoints to obtain the final model \cite{sys-amu-wmt16}.

 At inference time we decode with beam size 4 using SGNMT \cite{stahlberg2017sgnmt}. For BI we use 2-gram KENLM models \cite{heafield2011kenlm} trained on the source training data for each domain. For validation results we report cased BLEU scores with
SacreBLEU \cite{post2018call}\footnote{SacreBLEU signature: \texttt{BLEU+case.mixed\\+numrefs.1+smooth.exp+tok.13a+version.1.3.2}}; test results use case-insensitive BLEU.
\begin{figure*}[ht]
\centering
\small
\includegraphics[width=\linewidth]{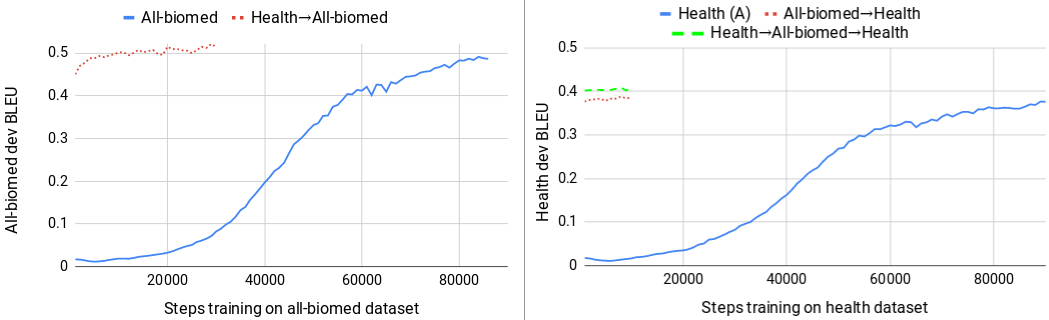}
\caption{Transfer learning for es2en domains. Left: standard transfer learning improves performance from a smaller (health) to a larger (all-biomed) domain. Right: returning to the original domain after transfer learning provides further gains on health.}
\label{fig:transfer}
\end{figure*}
\begin{table*}[ht]
\centering
\small
\begin{tabular}{|l|ccc|ccc|}

\hline
\textbf{Transfer learning schedule}  & \multicolumn{3}{c|}{\textbf{es2en}}  & \multicolumn{3}{c|}{\textbf{en2es}} \\
     &   \textbf{Khresmoi} & \textbf{Health} & \textbf{Bio} &   \textbf{Khresmoi} & \textbf{Health} & \textbf{Bio}   \\
     
     \hline
    Health &  45.1 & 35.7 & 34.0 & 41.2 & 34.7 & 36.1 \\
    All-biomed  & 49.8 & 35.4 & 35.7 & 43.4 & 33.9 & 37.5 \\
    All-biomed  $\rightarrow$ Health &48.9 & 36.4& 35.9 &  43.0 & 35.2 & 38.0 \\
    All-biomed  $\rightarrow$ Bio & 48.0 & 34.6 & 37.2&  43.2 & 34.1 & 40.5 \\
    \hline
    Health $\rightarrow$ All-biomed &  \textbf{52.1} & 36.7 &37.0 & 44.2 &35.0  &39.0   \\
    Health $\rightarrow$ All-biomed $\rightarrow$ Health   & 51.1 & \textbf{37.0}  & 37.2 &  44.0&\textbf{36.3} &39.5 \\
    Health $\rightarrow$ All-biomed $\rightarrow$ Bio  & 50.6 & 36.0 &\textbf{38.0} & \textbf{45.2} & 35.3 &\textbf{41.3} \\
\hline
\end{tabular}
\caption{Validation BLEU for English-Spanish models with transfer learning. We use the final three models in our submission.}\label{tab:transfer-results}
\end{table*}

\begin{table*}[ht]
\centering
\small
\begin{tabular}{|l|cccc|cccc|}
\hline
  & \multicolumn{4}{c|}{\textbf{es2en}}  & \multicolumn{4}{c|}{\textbf{en2es}} \\
     &   \textbf{Khresmoi} & \textbf{Health} & \textbf{Bio} &  \textbf{Test} & \textbf{Khresmoi} & \textbf{Health} & \textbf{Bio}  &  \textbf{Test} \\ \hline
    Health $\rightarrow$ All-biomed  &  52.1 & 36.7 &37.0& 42.4 & 44.2 &35.0  &39.0 &  44.9 \\
    Health $\rightarrow$ All-biomed $\rightarrow$ Health   & 51.1 & 37.0 & 37.2 & -&  44.0&36.3&39.5 &-\\
    Health $\rightarrow$ All-biomed $\rightarrow$ Bio   & 50.6 & 36.0 &38.0 & -& 45.2& 35.3 &41.3&  -\\
    \hline
    Uniform ensemble  &\textbf{52.2} &36.9 & 37.9& \textbf{43.0} & \textbf{45.1}& 35.6& 40.2 & 45.4\\
BI ensemble  ($\alpha$=0.5) & 52.1 & \textbf{37.0} &\textbf{38.1} & 42.9 & 44.5  & \textbf{35.7} &\textbf{41.2} &\textbf{45.6} \\
\hline
\end{tabular}
\caption{Validation and test BLEU for models used in English-Spanish language pair submissions.}\label{tab:submission-results-esen}
\end{table*}

\begin{table*}[ht]
\centering
\small
\begin{tabular}{|l|ccc|ccc|}
\hline
 & \multicolumn{3}{c|}{\textbf{de2en}}  & \multicolumn{3}{c|}{\textbf{en2de}} \\
     &   \textbf{Khresmoi} & \textbf{Cochrane} &  \textbf{Test}&   \textbf{Khresmoi} & \textbf{Cochrane} &  \textbf{Test}  \\ \hline
    News & 43.8  & 46.8  & -&  30.4 & 40.7  & -\\
    News $\rightarrow$ All-biomed & 44.5 & 47.6& 27.4& 31.1& 39.5 & 26.5\\ 
    \hline
    Uniform ensemble  & 45.3  & 48.4& \textbf{28.6} & \textbf{32.6} &    42.9 & \textbf{27.2}\\
BI ensemble ($\alpha$=0.5) & \textbf{45.4}  & \textbf{48.8} & 28.5& 32.4    &  \textbf{43.1} & 26.4\\
\hline
\end{tabular}
\caption{Validation and test BLEU for models used in English-German language pair submissions.}\label{tab:submission-results-ende}
\end{table*}

\begin{table}
\centering
\small
\begin{tabular}{|c|cccc|} \hline
 & \textbf{es2en} & \textbf{en2es} & \textbf{de2en}& \textbf{en2de}\\
 \hline
 Uniform  & \textbf{43.2} & 45.3& 28.3 & 25.9 \\
 BI  ($\alpha$=0.5) &43.0 & \textbf{45.5} &28.2 &25.2 \\
 BI  ($\alpha$=0.1) & \textbf{43.2} & \textbf{45.5} & \textbf{28.5} & \textbf{26.0}\\
 \hline
\end{tabular}
\caption{Comparing uniform ensembles and BI with varying smoothing factor on the WMT19 test data. Small deviations from official test scores on submitted runs are due to tokenization differences.  $\alpha=0.5$ was chosen for submission based on results on available development data.}\label{tab:bi01}
\end{table}
\subsection{Results}

Our first experiments involve iterative transfer learning in es2en and en2es to obtain models on three separate domains for the remaining evaluation. We use health, a relatively clean and small dataset, as the initial domain to train from scratch. Once converged, we use this to initialise training on the larger, noiser all-biomed corpus. When the all-biomed model has converged, we use it to initialise training on the health data and bio data for stronger models on those domains. Figure \ref{fig:transfer} shows the training progression for the health and all-biomed models, as well as the standard transfer learning case where we train on all-biomed from scratch. 

Table \ref{tab:transfer-results} gives single model validation scores for es2en and en2es models with standard and iterative transfer learning. We find that the all-biomed domain gains 1-2 BLEU points from transfer learning. Moreover, the health domain gains on all domains from iterative transfer learning relative to training from scratch and relative to standard transfer learning(All-biomed $\rightarrow$ Health), despite being trained twice to convergence on health.

We submitted three runs to the WMT19 biomedical task for each language pair: the best single all-biomed model, a uniform ensemble of models on two en-de and three es-en domains, and an ensemble with Bayesian Interpolation. Tables \ref{tab:submission-results-esen} and \ref{tab:submission-results-ende} give validation and test scores.

We find that a uniform multi-domain ensemble performs well, giving 0.5-1.2 BLEU improvement on the test set over strong single models. We see small gains from using BI with ensembles on most validation sets, but only on en2es test. 

Following test result release, we noted that, in general, we could predict BI ($\alpha=0.5$) performance by comparing the uniform ensemble  with the oracle model performing best on each validation domain. For en2es uniform ensembling underperforms the health and bio oracle models on their validation sets, and the uniform ensemble slightly underperforms BI on the test data. For en2de, by contrast, uniform ensembling is consistently better than oracles on the dev sets, and outperforms BI on the test data. For de2en and es2en, uniform ensembling performs similarly to the oracles, and performs similarly to BI. 

From this, we hypothesise that BI ($\alpha=0.5$) has a tendency to converge to a single model. This is effective when single models perform well (en2es) but ineffective if the uniform ensemble is predictably better than any single model (en2de). Consequently in Table \ref{tab:bi01} we experiment with BI ($\alpha=0.1$). In this case BI matches or out-performs the uniform ensemble. Notably, for en2es, where  BI ($\alpha=0.5$) performed well,  taking $\alpha=0.1$ does not harm performance. 

\section{Conclusions}
Our WMT19 Biomedical submission covers the English-German and English-Spanish language pairs, achieving the best submitted results on both directions of English-Spanish. We use transfer learning iteratively to train single models which perform well on related but distinct domains, and show further gains from multi-domain ensembles. We explore Bayesian Interpolation for multi-domain ensemble weighting, and find that a strongly smoothed case gives small gains over uniform ensembles.

\section*{Acknowledgments}
This work was supported by EPSRC grant EP/L027623/1 and has been performed using resources provided by the Cambridge Tier-2 system operated by the University of Cambridge Research Computing Service\footnote{\url{http://www.hpc.cam.ac.uk}} funded by EPSRC Tier-2 capital grant EP/P020259/1.

\bibliographystyle{acl_natbib}
\bibliography{refs}

\end{document}